\documentclass[10pt,a4paper]{article}

\usepackage{enumitem}

\usepackage{epstopdf}
\usepackage{psfrag}
\usepackage{graphics}% Others may be landscape, longtable, algorithm, algorithmic, etc.
\usepackage{lineno} % This package together with lineno.sty numbers every line. Makes it easy for edditing.
\usepackage{graphicx} % For including images
\usepackage{subcaption} % For subfigures

\usepackage{amsthm}
\usepackage{booktabs} % For professional-quality tables
\usepackage{float} % For the [H] option in tables and figures

\usepackage{array} % For extended column formatting options
\usepackage{float} % For the [H] option in tables and figures

\usepackage{titlesec}
\titleformat{\subsection}
  {\normalfont\bfseries}{\thesubsection}{1em}{}
\usepackage{enumitem}
\usepackage[utf8]{inputenc}
\usepackage[T1]{fontenc}
\usepackage{amsmath}
\usepackage{amsthm}
\usepackage{amsfonts}
\usepackage{amssymb}
\usepackage{makeidx}
\usepackage{graphicx}
\usepackage{algorithm2e}
\usepackage{lmodern}
\usepackage{mathtools}
\usepackage[utf8]{inputenc}
\usepackage[numbers]{natbib}
\usepackage{url}
\usepackage{parskip}

\usepackage{xcolor}
\usepackage[left=2cm,right=2cm,top=1cm,bottom=2cm]{geometry}

 \author{ Prudence Djagba\thanks{Michigan State University, USA. 
 		\href{mailto:callixte.ndizihiwe@aims.ac.rw}{djagbapr@msu.edu }}
 \and 
 Callixte Ndizihiwe \thanks{AIMS Rwanda, Kigali. 
		\href{mailto:prudence.djagba@aims.ac.rw}{callixte.ndizihiwe@aims.ac.rw}} 
	}
\title{ Pricing American Options using Machine Learning
Algorithms}
\date{}

\usepackage{hyperref}
\usepackage{cleveref}

\theoremstyle{remark}

\theoremstyle{definition}

\begin{document}
	\maketitle

\begin{abstract} 
		This study investigates the application of machine learning algorithms, particularly in the context of pricing American options using Monte Carlo simulations. Traditional models, such as the Black-Scholes-Merton framework, often fail to adequately address the complexities of American options, which include the ability for early exercise and non-linear payoff structures. By leveraging Monte Carlo methods in conjunction Least Square Method machine learning was used. This research aims to improve the accuracy and efficiency of option pricing. The study evaluates several machine learning models, including neural networks and decision trees, highlighting their potential to outperform traditional approaches. The results from applying machine learning algorithm in LSM indicate that integrating machine learning with Monte Carlo simulations can enhance pricing accuracy and provide more robust predictions, offering significant insights into quantitative finance by merging classical financial theories with modern computational techniques. The dataset was split into features and the target variable representing bid prices, with an 80-20 train-validation split. LSTM and GRU models were constructed using TensorFlow's Keras API, each with four hidden layers of 200 neurons and an output layer for bid price prediction, optimized with the Adam optimizer and MSE loss function. The GRU model outperformed the LSTM model across all evaluated metrics, demonstrating lower mean absolute error, mean squared error, and root mean squared error, along with greater stability and efficiency in training.
  
   \textbf{keywords:} Machine Learning, American Options, Monte Carlo Simulations, Least Square Method, Neural Networks,  Option Pricing,  LSTM, GRU.
  \end{abstract}
 
		%MSC: 16Y30,12K05
  \newpage
		\section{Introduction}
The pricing of American options is a complex task due to their early exercise feature. This study explores the use of machine learning models to enhance the Least Squares Monte Carlo (LSM) method for pricing American options. An option is a financial derivative that gives the buyer the right to buy or sell an underlying asset upon paying a premium. A call option gives the buyer the right to buy an asset while a put option gives the buyer the right to sell an asset \citep{li2023deep}. Option pricing is a crucial aspect of financial markets and has undergone extensive study and development. Accurate options valuation is essential for investors, traders, and financial institutions to make informed decisions and effectively manage risk. In recent years, advances in computational techniques and the availability of large datasets have paved the way for applying machine learning algorithms in option pricing. Machine learning techniques, particularly deep learning, can enhance option pricing models by capturing complex patterns and relationships in market data \citep{black1973pricing}. Unlike traditional models, which rely on predetermined formulas and assumptions, machine learning algorithms can adapt to changing market conditions and incorporate a wider range of input variables, leading to more accurate and robust pricing predictions \citep{black1973pricing}. By leveraging historical market data, machine learning algorithms can learn from past pricing dynamics and adapt to changing market conditions, thereby enhancing option pricing accuracy. The pricing of financial derivatives, particularly options, has been the subject of significant research and development in the field of quantitative finance. Traditional option pricing models, such as the Black-Scholes model \cite{merton1998applications}, have provided valuable insights into the valuation of European options. However, pricing American options, which allow for early exercise, presents unique challenges due to their non-linear payoff structure.
Accurate option pricing is crucial for the stability and efficiency of financial markets. However, investors, traders, and financial institutions often face challenges in determining precise prices for financial derivatives. These challenges can result in suboptimal decision-making and increased financial risk. The pricing of American options is particularly problematic due to their allowance for early exercise, which adds a layer of complexity that traditional models struggle to address. This complexity necessitates the exploration of alternative approaches that can provide more accurate and reliable pricing. To address these challenges, this research aims to employ Least Square Monte Carlo (LSM) methods combined with machine learning models. By leveraging these advanced techniques, we aim to better understand complex market patterns and improve the accuracy of American option pricing. The study of machine learning applications in option pricing is significant for several reasons. Firstly, it has the potential to enhance the development of financial markets by providing more accurate pricing models. Improved pricing models can lead to better decision-making and risk management for financial professionals, thereby contributing to the overall stability and efficiency of the financial system. Secondly, this research addresses the limitations of traditional pricing models, such as their reliance on predetermined assumptions and inability to adapt to changing market conditions. By integrating machine learning with Monte Carlo simulations, this study aims to develop models that are more flexible and capable of capturing the complexities of financial markets. Furthermore, the findings of this study could provide valuable insights into the factors that influence option prices, thereby advancing our understanding of financial markets and improving the tools available for quantitative finance.

		\newpage

		%Especiellay for $|R|=2$ the sequence are known as Bell numbers, which count the total number of partitions.
		
		\section{Literature Review}
		Previous studies have applied various techniques to price American options. The LSM method, introduced by Longstaff and Schwartz (2001), is widely used due to its flexibility and accuracy. Recent advancements in machine learning have shown promise in improving the estimation of continuation values in the LSM algorithm.
		\subsection{Risk Neutral Pricing}

Risk neutral pricing is a fundamental concept in financial mathematics used to evaluate the fair value of derivatives contracts, such as options, in a manner that accounts for risk without introducing arbitrage opportunities \cite{bingham2013risk}. This approach relies on the assumption that investors are indifferent to risk when pricing financial assets, allowing for a simplified valuation framework \cite{bingham2013risk}.

Considering a model economy with risky assets \(n_S \), risk-free assets \(\beta \), and risky assets \(S_i \). An equation for a differential is followed by the risk-free asset \(\beta \): 

\begin{equation}
d\beta(t) = \beta(t)r(t)dt,
\end{equation}

where the risk-free interest rate is denoted by \( r(t) \). The stochastic differential equation governing the hazardous assets \( S_i \) in the real-world measure \( P \) is as follows:

\begin{equation}
dS_i(t) = S_i(t)(\mu_i(t)dt + \sigma_i(t)dW^P_i),
\end{equation}

where \(dW^P_i \) is a \(n_S \)-dimensional standard Brownian motion, \( \sigma_i(t) \) represents the asset's volatility, and \( \mu_i(t) \) represents the asset-dependent drift term \cite{mckean1963brownian}.

The price \( V(S(t_0), t_0) \) of a European type contract in a complete market is determined by the expected value of the future price \( V(S(t_1), t_1) \) in relation to the risk-neutral measure \( Q \) \cite{bingham2013risk}, given as follows:

\begin{equation}
V(S(t_0), t_0) = E^Q \left[ \frac{V(S(t_1), t_1)}{\beta(t_0) \beta(t_1)} \right],
\label{2.2.3}
\end{equation}

The equation (\ref{2.2.3}) ensures that there are no arbitrage opportunities in the market. The measure \( Q \) is unique under the assumption of an arbitrage-free market. In the risk-neutral measure \( Q \), all the drift terms of the risky assets \( S_i \) are given by the risk-free interest rate \( r(t) \):
\begin{equation}
dS_i(t) = S_i(t) \left( r(t)dt + \sigma_i dW^Q_i(t) \right), \label{2.4}
\end{equation}

where \( \sigma_i \) is the asset's volatility, and \( W^Q_i(t) \) is a standard Brownian motion in the \( Q \) measure.
The relationship between the Brownian motions \( W^Q_i(t) \) and \( W^P_i(t) \) is given by:

\begin{equation}
dW^Q_i(t) = dW^P_i(t) + \nu_i(t) dt,
\end{equation}

where \( \nu_i(t) \) satisfies \( \mu(t) = r(t) + \sigma_i \nu_i(t) \). Notably, the volatilities \( \sigma_i \) remain the same under this change of measure, allowing for the estimation of \( \sigma_i \) from real-world observations of the asset price processes \( S_i \).

It is instantly evident from combining the differential equations for the risky and risk-free assets that the ratio \( \frac{S_i}{\beta} \) is a drift-free process \cite{lintner1975valuation}:

\begin{equation}
d\left( \frac{S_i}{\beta(t)} \right) = \left( \frac{S_i}{\beta(t)} \right) \sigma_i dW^Q_i(t).
\end{equation}

% This property, along with the expectation equation earlier, forms the basis of risk-neutral pricing, enabling the calculation of derivative prices under the assumption of risk neutrality. 

Risk-neutral pricing provides a powerful and wide framework for valuing financial derivatives, offering simplicity and tractability in complex market environments. By incorporating the principles of risk neutrality, financial analysts can derive fair prices for derivatives contracts, facilitating informed investment decisions and risk management strategies.
\subsection{ European options}
A European option is a form of options agreement where execution is restricted solely to its expiry date \cite{yoshida2003valuation}. This means that investors cannot exercise the option prematurely regardless of any fluctuations in the price of the underlying security such as a stock. The execution of the call or put option is only permitted on the date of the option's maturity.
 At time $t$, the cost of a European option is provided by

\begin{equation}
V(S(t), t) = \mathbb{E}^Q \left[ h(S(T), T) \exp \left( - \int_{t}^{T} r(s)ds \right) \right], \label{2.7}
\end{equation} 

where $h(S, T)$ is the payoff of the option at maturity, e.g.,
\begin{equation}
h(S, T) = \left\{
\begin{array}{ll}
\max[S - K, 0], & \text{for a call option}, \\
\max[K - S, 0], & \text{for a put option},
\end{array}
\right. 
\end{equation}
with $K$ the strike of the option.

\subsection{ American options}

Options with an extra right for the contract holder are known as American options. Anytime prior to or on the day of expiration, the option may be exercised. Due to this additional right, an American choice may be worth more than a European option. The European option will always have a higher payoff if it is exercised before it expires, so the American option can never be worth less than the European option. However, in certain situations, the extra right to exercise it early may allow for a higher payoff \cite{yoshida2003valuation}.

An American option holder may exercise their right to do so at any time up until and including maturity, unlike holders of European options. Because of this qualitative difference, in the American context as compared to the European case, the option holder has more rights. An American option's pricing must be at least equal to that of a comparable European option.

An American option holder must continually check the price of the underlying asset throughout the option's lifetime and determine if the option's price exceeds the instant payout they would get if they exercised the option at this particular moment.
 It can be shown that the price $V(S, t)$ of an American option is provided by
\begin{equation}
V(S(t), t) = \sup_{\tau\in[t,T]} \mathbb{E}^Q \left[ \exp \left( -\int_{t}^{\tau} r(s)ds \right) h(S(\tau)) \right],\label{e} 
\end{equation}

where the ideal stopping time supremum is attained 
$\tau^*$
\begin{equation}
\tau^* = \inf_{t\geq 0} \{ \tau: V(S(t), t) \leq h(S(t)) \}, 
\end{equation}
For the first time, the option's price is below what the holder would receive if they exercised it at this particular moment.
		% \paragraph{Acknowledgments}
		% We thank AIMS. 

  \section{Methods }
  The LSM method involves simulating paths of the underlying asset prices and performing a backward induction to estimate option values. In this study, machine learning models, including XGBoost, LightGBM, and logistic regression, are integrated into the LSM framework to estimate the continuation value more accurately.
  \subsection{Least-Squares Monte Carlo Method}

In the work \cite{boyle1977options}, Monte Carlo simulation is introduced into the financial domain and Monte Carlo techniques are used to price structured goods.
 These techniques provide an effective approximation of the option price, especially for multidimensional issues like derivatives with numerous underlying assets. European options in particular benefit from this, but American options can also be priced rather effectively. The following is the fundamental notion underlying the Monte Carlo Simulations technique for an option with payoff $h$. The price of the derivative is determined by explicitly computing the expected value of the discounted payoff, as it is in equation (\ref{2.7}), using these paths to create a (large) sample of random processes of the equation (\ref{2.4}) for the underlying stochastic processes.

 \subsection{Simulating random paths}

Solving the following system of coupled stochastic differential equations is necessary in order to simulate the paths of the underlying assets.

 \begin{equation}
 dS_i(t) = S_i(t) \left( r(t)dt + \sigma_i dW_i^Q(t) \right),
 \end{equation}
where $r(t)$ is the deterministic interest rate and $\mathbf{W}$ is a $n$-dimensional $Q$-Brownian motion with correlation matrix $\rho_{ij}$. 

 This can be accomplished most conveniently with the Euler-Maruyama approach. \cite{boyle1977options} has a thorough introduction. For a time mesh $t_j = t_0 + jdt$, $j = 1, \ldots, \text{nstep}$, step size $dt$, essentially, simulating random pathways means computing

 \begin{equation}
 S_i(t + dt) = S_i(t) \left( 1 + r(t)dt + \sqrt{dt} \sum_j B_{ij} Z_j(t) \right),
 \end{equation}
 where $Z_j(t)$ are independent standard normal random variables, and $B_{ij}$ is derived from the Cholesky decomposition of the correlation matrix $\rho_{ij}$.

% For each time step $\Delta t$ or in matrix form
% \begin{equation}
% \tilde{S}(t + \Delta t) = \tilde{S}(t) \left( \mathbf{1} + r(t)\Delta t\mathbf{1} + \sqrt{\Delta t}B \cdot \tilde{Z} \right),
% \end{equation}
% where the multiplication is component-wise and $Z_j(t)$ are uncorrelated random numbers.
% $B$ is a matrix that factorizes the covariance matrix $\Sigma = B B^T$ and can be obtained via
% Cholesky decomposition. The covariance matrix $\Sigma_{ij} = \rho_{ij}\sigma_i\sigma_j$ encodes the correlation $\rho_{ij}$
% and variances $\sigma_i$ of the assets (no sum over $i, j$).

\subsection{LSM for European Options}

\begin{figure}
    \centering
    \includegraphics[width=0.7\textwidth]{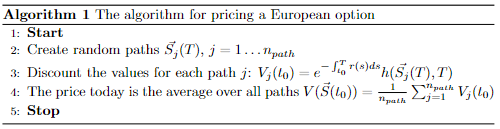}
    \caption{The algorithm for pricing a European option is a straightforward implementation of equation (\ref{2.7}) \cite{fang2009novel}.}
    \label{Figure 3.1:}
\end{figure}

With a European option, the Black-Scholes model may precisely sample the end-point of the pathways, which is the sole point that matters for the payout and doesn't require Euler-Maruyama time-stepping. We can calculate the value of the payout at maturity for each path after generating an ensemble of random values for the underlying assets' value at maturity, $S^j_i(T)$. The current option price can be obtained by averaging the discounted payoffs of all simulated paths, as illustrated in Figure \ref{Figure 3.1:}. The standard deviation can be used to calculate the degree of price uncertainty.

% \subsection{American Options}

% American options present a more complicated scenario compared to European options. At each time
% step, the holder of an American option must decide whether to exercise the option or continue holding it. Essentially, if the value of keeping the American option is less than the immediate payoff, the rational decision would be to exercise the option right away. 
\subsection{The Least Squares Monte Carlo (LSM) Algorithm for American Options}
To price an American option using Monte Carlo simulation for \( n \) underlying assets, a backward iteration algorithm is employed. The Least Squares Monte Carlo (LSM) algorithm is a method for pricing options by simulating potential future paths of the underlying asset's price and recursively working backward through time. It begins by generating random paths for the asset's price and setting the option's payoff at maturity based on the payoff function. Then, starting at maturity, it discounts future payoffs, performs regression analysis to estimate continuation values, and compares them to immediate exercise values to determine whether to exercise the option. This process iterates backward through time until the present. At the final time step, option values are discounted back to the present, and the option price is computed by averaging over all paths. LSM is particularly effective for pricing American-style options due to its ability to account for early exercise opportunities through regression analysis, making it a versatile and accurate approach for pricing complex derivatives.

Expand the continuation value \( c(\mathbf{S}) \) (as a function of the underlying asset price) at each time step \( t_i \) in terms of a function basis \( \psi_j \).
Be aware that it costs a lot of money to compute an option's continuation value.
Consequently, using a least squares regression to approximate the continuation value at each time across all pathways,

\begin{equation}
c(\mathbf{x}, t_i) = \mathbb{E} \left[  V(S(t_i+1)) \mid \mathbf{S}(t_i) = \mathbf{x} \right] = \sum_{k=0}^{n_{\text{order}}} \beta_k \psi_k(\mathbf{x}), \label{3.4}
\end{equation}
where the expansion coefficients \( \beta_k \) are obtained by a least squares fit to the (discounted) values of the option at the next time step:
\begin{equation}
\beta = \left( B_{\psi\psi} \right)^{-1} B V_\psi,  \label{3.5}
\end{equation}
where \( B_{\psi\psi} \), \( B V_\psi \) at time step \( t_i \) are given by
\begin{align}
(B_{\psi\psi})_{\ell} &= \mathbb{E} \left[ V(S(t_i+1)) \psi_{\ell}(S(t_i)) \right],  \\
(B_{\psi\psi})_{k\ell} &= \mathbb{E} \left[ \psi_k(S(t_i)) \psi_{\ell}(S(t_i)) \right], 
\end{align}
and the expectation is over the ensemble of paths.

\begin{figure}[H]
    \centering
    \includegraphics[width=1 \textwidth]{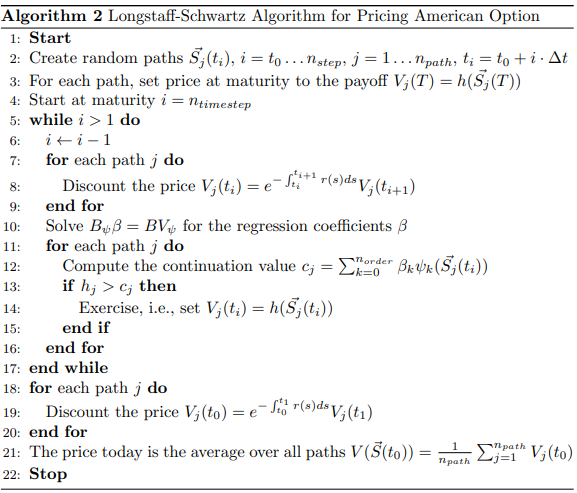}
    \caption{The algorithm for pricing American option }
    \label{AL}
\end{figure}

Unlike when the price is chosen to replicate the early exercise decision

\[ V_j(t_i) = h(S_j(t_i)),  \]
for all paths \( j \) where \( h(S_j(t_i)) > e^{- \int_{t_i}^{t_{i+1}} r(s)ds} V_j(t_{i+1}) \) in step 6b in Figure \ref{AL}, in some papers like \cite{stentoft2004convergence} assume that the non exercised value is the continuation value
\[ V_j(t_i) = \max \{ h(c(S_j(t_i))), h(S_j(t_i)) \}. \]
Its disadvantage is that the sampling error is compounded by the difference between \( c \) and \( h \),

where

\begin{itemize}
    \item \( t_i \): Represents each time step in the option pricing process.
    \item \( c(\mathbf{x}, t_i) \): The continuation value of the option at time \( t_i \), where \( \mathbf{x} \) denotes the underlying asset price.
    \item \( h \): The payoff function of the option.
    \item \( V(S(t_i+1)) \): The value of the option at the next time step \( t_{i+1} \) given the asset price \( S(t_i+1) \).
    \item \( \psi_j \): A function basis used for expansion.
    \item \( \beta_k \): Expansion coefficients obtained through a least squares fit.
    \item \( n_{\text{order}} \): The order of the expansion.
    \item \( B_{\psi\psi} \): The matrix of expectations of the product of basis functions.
    \item \( B V_\psi \): The vector of expectations of the product of the option value and basis functions.
\end{itemize}

Figure \ref{AL} shows the American option pricing and detailing algorithm of the Longstaff-Schwartz method for using least squares regression in step 6 to calculate the continuation value c. The solution of equation (\ref{3.5}) to obtain equation (\ref{3.4}) is explained in detail in steps 6a and 6b in the purple boxes \cite{lin2021american}.

\subsection{Machine Learning Methods based on LSM}

This section details how various machine learning models can be integrated into the Least Squares Monte Carlo (LSM) algorithm to enhance the pricing of American options. The machine learning models discussed include XGBoost, LightGBM, logistic regression, k-nearest neighbors (kNN), decision tree, and random forest.

In the  LSM algorithm in Figure (\ref{AL}) for pricing American options, machine learning models are primarily involved in Step 6a, where the continuation value is estimated through regression. Traditionally, this step uses linear regression to estimate the relationship between the current state variables and the future payoffs. However, integrating machine learning models such as XGBoost, LightGBM, logistic regression, k-nearest neighbors (kNN), decision trees, and random forests can significantly enhance this process. These models are trained on the simulated paths and their corresponding discounted payoffs \(V_j(t_{i+1})\), allowing them to capture complex, non-linear relationships in the data. By doing so, they can provide more accurate predictions of the continuation value \(c_j\) for each path \(j\). For example, XGBoost and LightGBM are gradient-boosting models that can handle large datasets with intricate interactions, while decision trees and random forests can model non-linear relationships effectively.

Once the machine learning model is trained, it replaces the traditional linear regression formula \(B_{VV}\beta = B_{Vy}\) used to calculate the regression coefficients \(\beta\). In this enhanced approach, the model predicts the continuation value \(c_j\) for each path based on the state variables \(S_j(t_i)\). These predicted continuation values are then used to determine whether to exercise the option or to continue holding it (Step 7). If the payoff from exercising the option is greater than the predicted continuation value, the option is exercised; otherwise, it is not. By leveraging machine learning models in this critical step, the LSM algorithm can achieve more accurate and robust option pricing, as these models can generalize better to complex and high-dimensional state spaces than traditional linear regression.
\subsection{Recurrent Neural Networks (RNNs)}

Inspired by the architecture and operation of the human brain, neural networks (NNs) are a fundamental idea in machine learning. NNs are fundamentally made up of linked nodes arranged in layers. Data is received by input layers, information is processed by hidden levels, and output layers generate output. The capacity of NNs to learn from data and modify internal parameters (weights) during training to maximize performance is what gives them their strength \cite{schmidt2019recurrent}.

RNNs are a specific type of NN made to work with sequential data. They provide the idea of memory, which allows the network to remember data from earlier inputs. For jobs like pricing American options, where historical prices and market conditions might affect future decisions, this memory is essential \cite{schmidt2019recurrent}.

\text{Detailed working principals of Recurrent Neural Networks (RNNs) are presented as follows:  }

\begin{itemize}
    \item \textbf{Sequential Methodology:} RNNs, in contrast to conventional neural networks, are made to handle data sequences. They accomplish this by sequentially accepting inputs one at a time.
    
    \item \textbf{Repeated Relationships:} An RNN's recurrent connections are its primary characteristic. The network can maintain some sort of "memory" thanks to these links. The RNN processes the current input and a "hidden state" from the previous phase at each step in a sequence. Information gleaned from earlier inputs is contained in this hidden state.

    \item \textbf{Secret State:} At every time step \( t \), the hidden state \( h_t \) is updated using the prior hidden state \( h_{t-1} \) as well as the new input \( x_t \). In mathematics, this is commonly expressed as:

    \[
    h_t = \phi(W_{hx} x_t + W_{hh} h_{t-1} + b_h),
    \]
    where \( \phi \) is a non-linear activation function, \( W_{hx} \) and \( W_{hh} \) are weight matrices, and \( b_h \) is a bias vector.
    
    \item \textbf{Compared Weights:} All time steps in an RNN use the same weights, or parameters. This increases the efficiency of the model and lowers the number of parameters since the same weights are applied to each input in the sequence.

\end{itemize}

\subsection{Long Short-Term Memory (LSTM).}

In the realm of Recurrent Neural Networks, Long Short-Term Memory (LSTM) networks are an advanced evolution designed to overcome the drawbacks of conventional RNNs, especially when addressing long-term dependencies \cite{liu2023pre}.
 Therefore, the detailed processes throughout Long Short-Term Memory are explained as follows: 
\begin{itemize}
   \item \textbf{Higher Level Memory Management:} The LSTM unit, a sophisticated memory cell, is the distinguishing characteristic of LSTM. This unit's distinct structure, which consists of several gates, allows it to retain information for lengthy periods of time.
    
    \item \textbf{Gating System:} Three different kinds of gates are included in LSTMs, and each is essential to the network's memory management.

\begin{itemize}[label=$\star$] 
      \item \textbf{Input Gate:} Shows which values from the input should be used to modify the memory. Mathematically, the input gate \( i_t \) is defined as:
        \[
        i_t = \sigma(W_{ix} x_t + W_{ih} h_{t-1} + b_i),
        \]
        where \( \sigma \) is the sigmoid activation function, and \( W_{ix} \), \( W_{ih} \), and \( b_i \) are the weights and biases for the input gate.

        \item \textbf{Forget Gate:} Decides what portions of the existing memory should be discarded. The forget gate \( f_t \) is given by:
        \[
        f_t = \sigma(W_{fx} x_t + W_{fh} h_{t-1} + b_f),
        \]
        where \( W_{fx} \), \( W_{fh} \), and \( b_f \) are the weights and biases for the forget gate.

        \item \textbf{Output Gate:} Controls the output flow of the memory content to the next layer in the network. The output gate \( o_t \) is represented as:
        \[
        o_t = \sigma(W_{ox} x_t + W_{oh} h_{t-1} + b_o),
        \]
        where \( W_{ox} \), \( W_{oh} \), and \( b_o \) are the weights and biases for the output gate.
    \end{itemize}

    \item \textbf{Cell State:} The cell state \( C_t \), which functions as a kind of conveyor belt running straight down the length of the network chain, is the fundamental component of LSTM. It guarantees that the network efficiently stores and retrieves significant long-term information while permitting information to flow essentially unaltered. The following updates the cell state:

    \[
    C_t = f_t * C_{t-1} + i_t * \tilde{C}_t,
    \]
    where \( \tilde{C}_t \) is the candidate cell state, calculated as:
    \[
    \tilde{C}_t = \tanh(W_{cx} x_t + W_{ch} h_{t-1} + b_c),
    \]
    and \( \tanh \) is the hyperbolic tangent activation function \cite{liu2023pre}.
\end{itemize}

 % \begin{itemize}
 %        \item \textbf{Input Gate:} Determines which values from the input should be used to modify the memory. Mathematically, the input gate \( i_t \) is defined as:
 %        \[
 %        i_t = \sigma(W_{ix} x_t + W_{ih} h_{t-1} + b_i),
 %        \]
 %        where \( \sigma \) is the sigmoid activation function, and \( W_{ix} \), \( W_{ih} \), and \( b_i \) are the weights and biases for the input gate.

 %        \item \textbf{Forget Gate:} Decides what portions of the existing memory should be discarded. The forget gate \( f_t \) is given by:
 %        \[
 %        f_t = \sigma(W_{fx} x_t + W_{fh} h_{t-1} + b_f),
 %        \]
 %        where \( W_{fx} \), \( W_{fh} \), and \( b_f \) are the weights and biases for the forget gate.

 %        \item \textbf{Output Gate:} Controls the output flow of the memory content to the next layer in the network. The output gate \( o_t \) is represented as:
 %        \[
 %        o_t = \sigma(W_{ox} x_t + W_{oh} h_{t-1} + b_o),
 %        \]
 %        where \( W_{ox} \), \( W_{oh} \), and \( b_o \) are the weights and biases for the output gate.
 %    \end{itemize}

\subsection{Gated Recurrent Unit (GRU)}

GRUs are a cutting-edge variant of recurrent neural networks that aim to enhance and streamline LSTM architecture. They provide a more efficient method of managing sequential data, and they work especially well in situations where long-term dependencies are essential \cite{chung2014empirical}.  Therefore, the detailed processes throughout the Gated Recurrent Unit are explained as follows:

\begin{itemize}
    \item \textbf{Architecture Simplified:} In terms of processing resources, the GRU is more efficient because to its simpler structure than the LSTM. Its lower gate count accounts for this efficiency.
    
     \item Two gates are used by GRUs:

    \begin{itemize}[label=$\star$] 
    \item \textbf{ Update Gate:} The degree to which data from the previous state should be transferred to the present state is determined by this gate. It combines the forget and input gates that are present in LSTMs. We define the update gate \(z_t \) as follows:  

        \[
        z_t = \sigma(W_{zx} x_t + W_{zh} h_{t-1} + b_z),
        \]
        where \( W_{zx} \), \( W_{zh} \), and \( b_z \) are the weights and biases for the update gate.

        \item \textbf{Reset Gate:} It basically lets the model select how much of the past is useful for the current prediction by deciding how much of the past to ignore. Given is the reset gate \(r_t \):
        \[
        r_t = \sigma(W_{rx} x_t + W_{rh} h_{t-1} + b_r),
        \]
        where \( W_{rx} \), \( W_{rh} \), and \( b_r \) are the weights and biases for the reset gate.
    \end{itemize}

    \item\textbf{No Separate State for Cells:} There is no distinct cell state in GRUs, in contrast to LSTMs. In doing so, they streamline the information flow and facilitate modeling and training by merging the cell state and concealed state into a single structure.
 The hidden state \( h_t \) in a GRU is updated as follows:
    \[
    h_t = (1 - z_t) * h_{t-1} + z_t * \tilde{h}_t,
    \]
    where \( \tilde{h}_t \) is the candidate hidden state, calculated as:
    \[
    \tilde{h}_t = \tanh(W_{hx} x_t + r_t * (W_{hh} h_{t-1}) + b_h).
    \]
\end{itemize}

\subsection{Description of Dataset}

The data source used in this experimental investigation is a collection of historical data on all symbols in the U.S equities markets from January to June 2013
(\href{ https://optiondata.org }{https://optiondata.org}).

The given dataset provides detailed information on options contracts, encompassing various attributes crucial for options trading analysis. Bellow are explanations of each column:

\begin{itemize}
    \item \textbf{Contract}: A unique identifier for each options contract, likely containing information about the underlying asset, expiration date, type (call or put), and strike price.
    \item \textbf{Underlying}: Indicates the underlying asset associated with the options contract.
    \item \textbf{Expiration}: The expiration date of the options contract.
    \item \textbf{Type}: Specifies whether the option is a call or a put.
    \item \textbf{Strike}: The strike price of the options contract.
    \item \textbf{Style}: Refers to the style of the options contract (e.g., American or European).
    \item \textbf{Bid}: The bid price of the options contract, representing the highest price a buyer is willing to pay.
    \item \textbf{Bid Size}: The size of the bid, indicating the quantity of contracts being bid for.
    \item \textbf{Ask}: The ask price of the options contract, representing the lowest price a seller is willing to accept.
    \item \textbf{Ask Size}: The size of the ask, indicating the quantity of contracts being offered.
    \item \textbf{Volume}: The trading volume of the options contract.
    \item \textbf{Open Interest}: The total number of outstanding options contracts.
    \item \textbf{Quote Date}: The date when the quote for the options contract was made.
    \item \textbf{Delta}: Delta measures the rate of change of the option's price in response to changes in the price of the underlying asset.
    \item \textbf{Gamma}: Gamma measures the rate of change in delta in response to changes in the price of the underlying asset.
    \item \textbf{Theta}: Theta measures the rate of decline in the value of the option over time.
    \item \textbf{Vega}: Vega measures the sensitivity of the option's price to changes in implied volatility.
    \item \textbf{Implied Volatility}: Implied volatility is the market's estimate of the future volatility of the underlying asset, as implied by the options prices.
\end{itemize}

In the analysis of our dataset, we focused on the numerical features to understand the relationships between them. To achieve this, we computed the correlation matrix, which quantifies the linear relationships between pairs of numerical variables. We visualized this correlation matrix using a heatmap, a powerful tool for identifying patterns and correlations within the data. The heatmap, annotated for clarity, uses a 'coolwarm' color palette to indicate the strength and direction of the correlations, with positive correlations shown in warm tones and negative correlations in cool tones. This visual representation helps in quickly identifying strong correlations, both positive and negative, among the numerical features, thereby providing insights that can guide further analysis and decision-making processes. The heatmap underscores the importance of certain variables and their interdependencies, which can be critical for predictive modeling and other statistical analyses.

\begin{figure}
    \centering
    \includegraphics[width=0.7\textwidth]{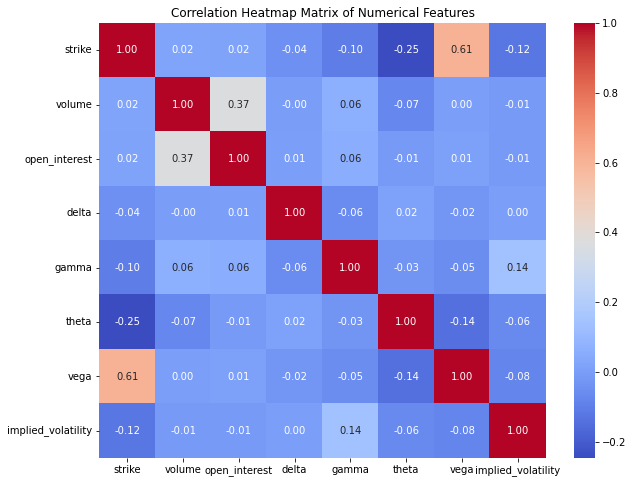}
    \caption{Correlation heatmap matrix amongst numerical features from the dataset.}
    \label{heet}
\end{figure}

The heatmap in Figure \ref{heet},  illustrates the correlation matrix of numerical features in the dataset, using the 'coolwarm' color palette to depict the strength and direction of correlations. Warm tones (\textcolor{red}{red}) indicate positive correlations, while cool tones (\textcolor{blue}{blue}) indicate negative correlations. The diagonal elements show a perfect correlation of 1, as each feature is perfectly correlated with itself. Notable observations include a strong positive correlation (0.61) between 'strike' and 'vega', and a moderate positive correlation (0.37) between 'volume' and 'open\_interest'. Conversely, 'strike' and 'theta' exhibit a moderate negative correlation (-0.25). Most feature pairs exhibit weak or no correlations, suggesting distinct underlying factors. This visualization aids in quickly identifying significant linear relationships, which is valuable for further analysis and decision-making.

\section{Results and analysis}
Table \ref{option_prices} and \ref{tab:option_prices} present the option prices and standard errors predicted by different machine learning models. The results indicate that models such as LightGBM and logistic regression outperform traditional linear regression in estimating continuation values.
\subsection{Result of LSM with Different Machine Learning Model} 

The results presented in Table \ref{option_prices} and \ref{tab:option_prices} offer valuable insights into the performance of LSM with machine learning algorithms in pricing American options.

\begin{table}[H]
\centering
\begin{tabular}{cccccccc}
\toprule
$S(0)$ & $\sigma$ & $T$ & \textbf{LGBM Price} & \textbf{LGBM Error} & \textbf{LR Price} & \textbf{LR Error} \\
\midrule
80 & 0.2 & 1 & 20.0972 & 0.0492 & 19.8687 & 0.0225 \\
80 & 0.2 & 2 & 20.4074 & 0.0654 & 19.8745 & 0.0323 \\
80 & 0.4 & 1 & 23.4062 & 0.0993 & 19.9867 & 0.0439 \\
80 & 0.4 & 2 & 26.2030 & 0.1298 & 20.3234 & 0.0574 \\
85 & 0.2 & 1 & 15.4515 & 0.0524 & 14.9698 & 0.0248 \\
85 & 0.2 & 2 & 16.0469 & 0.0667 & 15.0756 & 0.0341 \\
85 & 0.4 & 1 & 20.0282 & 0.1000 & 15.7400 & 0.0434 \\
85 & 0.4 & 2 & 23.2883 & 0.1230 & 16.6031 & 0.0546 \\
90 & 0.2 & 1 & 11.4459 & 0.0527 & 10.9653 & 0.0391 \\
90 & 0.2 & 2 & 12.5353 & 0.0643 & 11.5144 & 0.0436 \\
90 & 0.4 & 1 & 16.9559 & 0.0980 & 12.8807 & 0.0453 \\
90 & 0.4 & 2 & 20.7562 & 0.1226 & 14.2032 & 0.0537 \\
95 & 0.2 & 1 & 8.0805 & 0.0496 & 8.2777 & 0.0551 \\
95 & 0.2 & 2 & 9.6922 & 0.0630 & 9.3393 & 0.0561 \\
95 & 0.4 & 1 & 14.2657 & 0.0934 & 11.2833 & 0.0545 \\
95 & 0.4 & 2 & 18.6165 & 0.1194 & 12.8496 & 0.0574 \\
100 & 0.2 & 1 & 5.5364 & 0.0445 & 6.1403 & 0.0589 \\
100 & 0.2 & 2 & 7.1349 & 0.0578 & 7.3681 & 0.0617 \\
100 & 0.4 & 1 & 12.1516 & 0.0890 & 9.9857 & 0.0617 \\
100 & 0.4 & 2 & 16.3524 & 0.1166 & 11.7780 & 0.0637 \\
105 & 0.2 & 1 & 3.7669 & 0.0379 & 4.4493 & 0.0560 \\
105 & 0.2 & 2 & 5.3870 & 0.0516 & 5.9631 & 0.0617 \\
105 & 0.4 & 1 & 10.1960 & 0.0842 & 8.9259 & 0.0654 \\
105 & 0.4 & 2 & 14.6803 & 0.1116 & 10.8403 & 0.0676 \\
110 & 0.2 & 1 & 2.5966 & 0.0321 & 3.0940 & 0.0502 \\
110 & 0.2 & 2 & 4.0966 & 0.0455 & 4.7240 & 0.0596 \\
110 & 0.4 & 1 & 8.5805 & 0.0791 & 7.8738 & 0.0673 \\
110 & 0.4 & 2 & 13.2400 & 0.1093 & 10.0054 & 0.0716 \\
115 & 0.2 & 1 & 1.6877 & 0.0264 & 2.1591 & 0.0437 \\
115 & 0.2 & 2 & 3.1682 & 0.0413 & 3.6169 & 0.0552 \\
115 & 0.4 & 1 & 7.4315 & 0.0740 & 7.0224 & 0.0683 \\
115 & 0.4 & 2 & 11.7064 & 0.1039 & 9.3480 & 0.0728 \\
120 & 0.2 & 1 & 1.1187 & 0.0214 & 1.4166 & 0.0363 \\
120 & 0.2 & 2 & 2.3429 & 0.0356 & 2.8717 & 0.0510 \\
120 & 0.4 & 1 & 6.4239 & 0.0694 & 6.1176 & 0.0673 \\
120 & 0.4 & 2 & 10.5792 & 0.0994 & 8.4853 & 0.0744 \\
\bottomrule 
\end{tabular}
\caption{Option Prices and Standard Errors Predicted by Machine Learning Models}
\label{option_prices}
\end{table}

\begin{table}[H]
\centering
\begin{tabular}{p{0.04\linewidth}ccp{0.07\linewidth}p{0.07\linewidth}p{0.07\linewidth}p{0.07\linewidth}p{0.07\linewidth}p{0.07\linewidth}p{0.07\linewidth}p{0.07\linewidth}}
%\hline
%$S(0)$ & $\sigma$ & $T$ & \textbf{KNN Price} & \textbf{KNN StdErr} & \textbf{DT Price} & \textbf{DT StdErr} & \textbf{XGB Price} & \textbf{XGB StdErr} & \textbf{RF Price} & \textbf{RF StdErr} \\
%\hline
%\endfirsthead
\hline
$S(0)$ & $\sigma$ & $T$ & \textbf{KNN Price} & \textbf{KNN StdErr} & \textbf{DT Price} & \textbf{DT StdErr} & \textbf{XGB Price} & \textbf{XGB StdErr} & \textbf{RF Price} & \textbf{RF StdErr} \\
\hline
%\endhead
80 & 0.2 & 1 & 24.3728 & 0.0748 & 28.8665 & 0.0744 & 21.0675 & 0.0599 & 28.0579 & 0.0787 \\
80 & 0.2 & 2 & 25.3591 & 0.0964 & 30.9987 & 0.0931 & 21.5372 & 0.0785 & 30.0077 & 0.0975 \\
80 & 0.4 & 1 & 30.8710 & 0.1459 & 38.3669 & 0.1333 & 26.1217 & 0.1488 & 37.1172 & 0.1404 \\
80 & 0.4 & 2 & 34.6132 & 0.1770 & 42.8716 & 0.1580 & 29.0901 & 0.1802 & 41.6898 & 0.1723 \\
85 & 0.2 & 1 & 19.8393 & 0.0818 & 24.4773 & 0.0800 & 16.5987 & 0.0674 & 23.6552 & 0.0829 \\
85 & 0.2 & 2 & 21.0375 & 0.1013 & 26.8311 & 0.1000 & 17.4713 & 0.0893 & 25.9148 & 0.1053 \\
85 & 0.4 & 1 & 27.0881 & 0.1518 & 34.6294 & 0.1438 & 22.7602 & 0.1535 & 33.0799 & 0.1506 \\
85 & 0.4 & 2 & 31.2504 & 0.1815 & 40.0178 & 0.1696 & 26.4643 & 0.1826 & 38.3983 & 0.1821 \\
90 & 0.2 & 1 & 15.7616 & 0.0879 & 20.0486 & 0.0846 & 12.7415 & 0.0776 & 19.4280 & 0.0897 \\
90 & 0.2 & 2 & 17.3494 & 0.1080 & 22.6412 & 0.1052 & 13.8312 & 0.0977 & 21.6946 & 0.1138 \\
90 & 0.4 & 1 & 23.8535 & 0.1589 & 30.9616 & 0.1502 & 19.7064 & 0.1557 & 29.4104 & 0.1588 \\
90 & 0.4 & 2 & 28.3603 & 0.1888 & 36.9729 & 0.1802 & 24.1088 & 0.1840 & 34.8291 & 0.1922 \\
95 & 0.2 & 1 & 11.9447 & 0.0907 & 15.9148 & 0.0902 & 9.2868 & 0.0794 & 15.0424 & 0.0945 \\
95 & 0.2 & 2 & 13.7463 & 0.1090 & 18.8710 & 0.1132 & 10.8542 & 0.0970 & 17.6468 & 0.1178 \\
95 & 0.4 & 1 & 20.8209 & 0.1602 & 26.9232 & 0.1579 & 17.2811 & 0.1519 & 25.9173 & 0.1660 \\
95 & 0.4 & 2 & 25.8280 & 0.1917 & 33.6813 & 0.1899 & 21.9563 & 0.1863 & 32.4833 & 0.2008 \\
100 & 0.2 & 1 & 8.5302 & 0.0849 & 11.6348 & 0.0936 & 6.7459 & 0.0757 & 10.7567 & 0.0936 \\
100 & 0.2 & 2 & 11.0382 & 0.1074 & 14.9813 & 0.1161 & 8.8450 & 0.0950 & 14.1157 & 0.1197 \\
100 & 0.4 & 1 & 17.8818 & 0.1568 & 23.8793 & 0.1658 & 14.9276 & 0.1486 & 22.3133 & 0.1697 \\
100 & 0.4 & 2 & 23.6761 & 0.1902 & 30.4449 & 0.1951 & 19.9013 & 0.1845 & 28.8166 & 0.2036 \\
105 & 0.2 & 1 & 5.9984 & 0.0787 & 8.2016 & 0.0878 & 4.7206 & 0.0636 & 7.6624 & 0.0882 \\
105 & 0.2 & 2 & 8.5650 & 0.1008 & 11.7394 & 0.1151 & 6.8803 & 0.0857 & 11.0133 & 0.1171 \\
105 & 0.4 & 1 & 15.4037 & 0.1547 & 20.7673 & 0.1655 & 12.8568 & 0.1443 & 19.2972 & 0.1708 \\
105 & 0.4 & 2 & 21.0763 & 0.1909 & 27.4824 & 0.2002 & 18.0371 & 0.1788 & 26.0770 & 0.2066 \\
110 & 0.2 & 1 & 4.1100 & 0.0666 & 5.6554 & 0.0784 & 3.3785 & 0.0567 & 5.4294 & 0.0787 \\
110 & 0.2 & 2 & 6.7170 & 0.0919 & 9.1449 & 0.1082 & 5.5111 & 0.0796 & 8.4097 & 0.1085 \\
110 & 0.4 & 1 & 13.1887 & 0.1485 & 17.6178 & 0.1625 & 11.2735 & 0.1370 & 16.7252 & 0.1661 \\
110 & 0.4 & 2 & 19.1963 & 0.1891 & 25.3414 & 0.2022 & 16.4879 & 0.1730 & 23.5516 & 0.2021 \\
115 & 0.2 & 1 & 2.8160 & 0.0559 & 3.9509 & 0.0676 & 2.2665 & 0.0472 & 3.6378 & 0.0665 \\
115 & 0.2 & 2 & 4.9858 & 0.0828 & 7.2037 & 0.1004 & 4.1753 & 0.0710 & 6.6598 & 0.0991 \\
115 & 0.4 & 1 & 11.5542 & 0.1423 & 15.5455 & 0.1616 & 9.7780 & 0.1293 & 14.4947 & 0.1606 \\
115 & 0.4 & 2 & 17.2385 & 0.1822 & 22.7688 & 0.2002 & 14.5604 & 0.1694 & 21.7730 & 0.2028 \\
120 & 0.2 & 1 & 1.9654 & 0.0482 & 2.5620 & 0.0555 & 1.5674 & 0.0411 & 2.4316 & 0.0550 \\
120 & 0.2 & 2 & 3.8382 & 0.0726 & 5.3756 & 0.0895 & 3.1104 & 0.0617 & 5.0425 & 0.0872 \\
120 & 0.4 & 1 & 10.0550 & 0.1361 & 13.1948 & 0.1533 & 8.1217 & 0.1205 & 12.3349 & 0.1521 \\
120 & 0.4 & 2 & 15.6627 & 0.1798 & 20.4274 & 0.1973 & 13.3272 & 0.1629 & 19.5568 & 0.1986 \\
\hline 
\end{tabular}
\caption{Option Prices and Standard Errors Predicted by Machine Learning Models}
 \label{tab:option_prices}
\end{table}

To see the model performance we used the numerical example presented in work \cite{park2014parametric}.
In this numerical example, Table  \ref{option_prices} and \ref{tab:option_prices} report a range of numerical values for different parameter choices for different machine learning models. Throughout, we use \( K = 100 \), \( r = 0.04 \), \( T = 1 \), and take the same volatilities of both assets to be 0.2 or 0.4, with 10,000 paths and one basis function.

\subsection{Impact of Volatility and Time to Maturity}

In Table \ref{option_prices} and \ref{tab:option_prices}, we can observe the influence of volatility (\( \sigma \)) and time to maturity (\( T \)) on option prices. 

\begin{figure}[H]
     \centering
     \begin{subfigure}[b]{0.3\textwidth}
         \centering
    \includegraphics[width=0.8\linewidth]{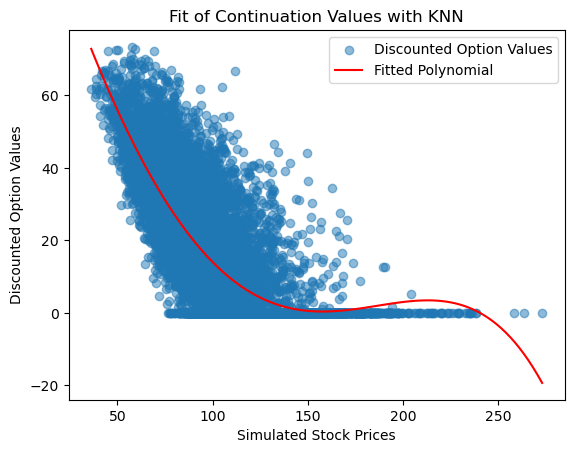}
        \caption{ KNN Model}
        \label{a}
     \end{subfigure}
     \hfill
     \begin{subfigure}[b]{0.3\textwidth}
         \centering
         \includegraphics[width=0.8\linewidth]{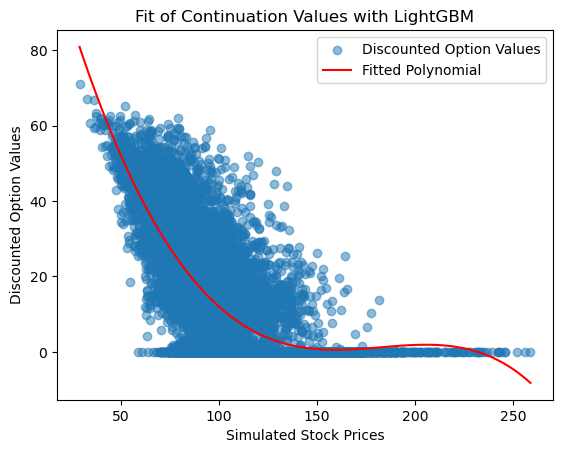}
         \caption{Light GBM Model}
       \label{b}
     \end{subfigure}
     \hfill
     \begin{subfigure}[b]{0.3\textwidth}
         \centering
        \includegraphics[width=0.8\linewidth]{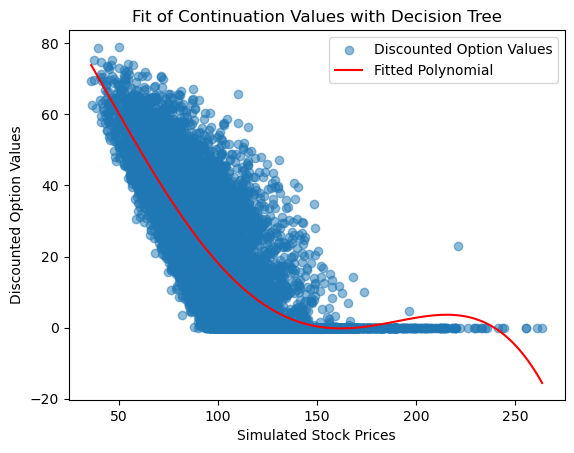}
        \caption{Decision tree Model}
        \label{c}
     \end{subfigure}
     
        \medskip
        \centering
     \begin{subfigure}[b]{0.3\textwidth}
         \centering         \includegraphics[width=0.8\linewidth]{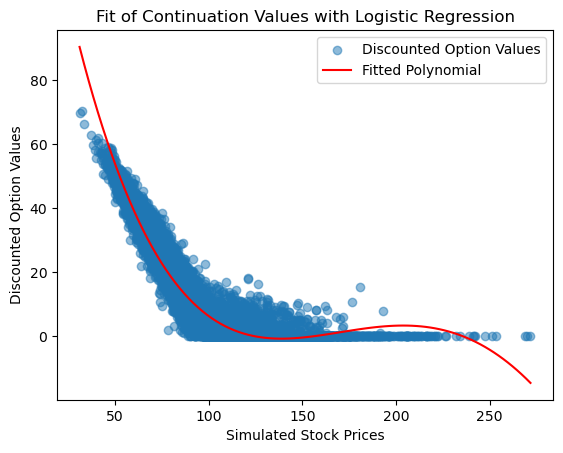}
         \caption{Logistic Rgression Model}
       \label{d}
     \end{subfigure}
     \hfill
     \begin{subfigure}[b]{0.3\textwidth}
         \centering
\includegraphics[width=0.8\linewidth]{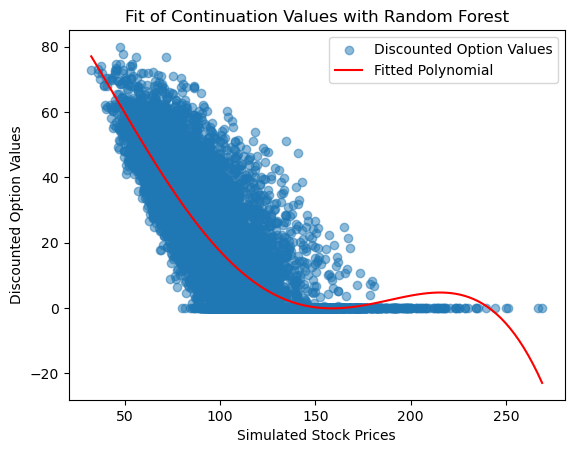}
        \caption{Random forest  Model}
        \label{l}
     \end{subfigure}
     \hfill
     \begin{subfigure}[b]{0.3\textwidth}
         \centering
         \includegraphics[width=0.8\linewidth]{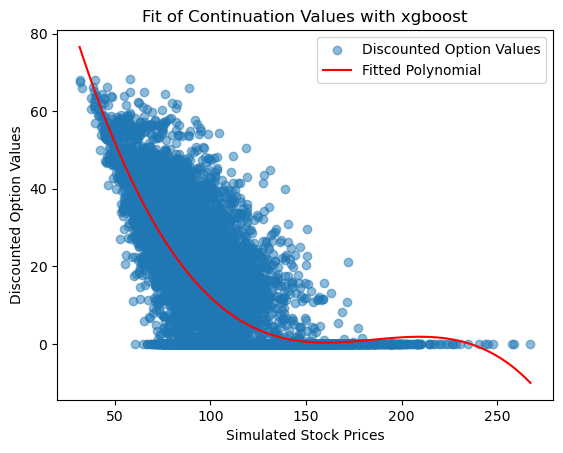}
         \caption{XGBoost Model}
       \label{f}
     \end{subfigure}
        \caption{Three simple graphs}
        \label{fig:MLMODELS}
\end{figure}

\begin{itemize}
  \item \textbf{Volatility (\( \sigma \)):} Higher volatility generally leads to higher option prices across all models. For instance, when \( S(0) = 80 \) and \( T = 1 \), the KNN price increases from 24.37 (when \( \sigma = 0.2 \)) to 30.87 (when \( \sigma = 0.4 \)). The standard errors tend to increase with higher volatility, reflecting the increased uncertainty and complexity in pricing under these conditions.
  
  \item \textbf{Time to Maturity (\( T \)):} Longer maturities also result in higher option prices. For example, with \( S(0) = 80 \) and \( \sigma = 0.2 \), the KNN price increases from 24.37 (when \( T = 1 \)) to 25.36 (when \( T = 2 \)). The standard errors typically increase with longer maturities, indicating greater variability in the predictions as the time horizon extends.
\end{itemize}
Figure \ref{fig:MLMODELS} provides a visual comparison of the fit of the six approaches. Each model here uses 5 basis functions (polynomials up to degree 5). The x-axis shows the simulated stock price at a step in time and the y-axis shows the discounted option value from the next step. Figure \ref{fig:MLMODELS}, we can see that polynomial fits for the following machine learning applied to LSM; KNN, Decision Tree, XGBoost, LightGBM, Logistic Regression, and Random Forest. Some of these methods provide a better visual fit for example here we can see that Logistic Regression provides a better visual fit.

\vspace{0.5cm}

Here let's compare the pricing of an example American put option with the following details:    $S_0 = 100$, $K = 100$, $T = 1.0$, $r = 0.02$, and $\sigma = 0.4$
 Monte Carlo simulation was performed with 25 time intervals and 10000 paths. Table \ref{tab:comparison} shows the results.

 \vspace{0.5cm}

\begin{table}[H]
    \centering
    \begin{tabular}{|c|c|c|c|c|c|c|}
        \hline
        \textbf{Method} & \textbf{ Option Value} & \textbf{Total Time (sec)}  \\
        \hline
        KNN & 18.8414 & 2.593  \\  
        \hline
        Decision Tree & 24.5571& 2.033  \\
        \hline
        XGBoost &  15.8242 & 6.689   \\
        \hline
        LightGBM & 15.6058 & 5.506  \\
        \hline
        Logistic Regression & 4.164 & 2.133  \\
        \hline
        Random Forest &23.312 &167.095   \\
        \hline
    \end{tabular}
    \caption{Comparison of American put option pricing and computational time using various methods}
    \label{tab:comparison}
\end{table}
From Table \ref{tab:comparison}, we can see computational times reveal that simpler models like Decision Trees and Logistic Regression are the quickest, with times of 2.033 and 2.133 seconds respectively, indicating their suitability for scenarios requiring rapid computations. KNN also performs moderately well at 2.593 seconds. In contrast, ensemble methods such as Random Forest take significantly longer, with a computational time of 167.095 seconds, reflecting their higher complexity and resource demands. Gradient boosting methods, XGBoost and LightGBM, offer a balance between speed and performance, with times of 6.689 and 5.506 seconds respectively, making them efficient yet powerful options. This highlights a trade-off between computational efficiency and model complexity, guiding the choice of method based on specific needs for speed versus predictive accuracy.

Below we are going to see the performance of various machine learning models used within the Long staff-Schwartz Method (LSM) framework for pricing American options. The models evaluated are K-Nearest Neighbors (KNN), Decision Tree, XGBoost, LightGBM, Logistic Regression, and Random Forest. Each model is analyzed based on the estimated option price, standard error, execution time, and classification performance metrics (confusion matrix, precision, recall, F1-score, and ROC AUC). Table \ref{t11} presents the results of six different models evaluated using several performance indicators for the in-time sample. These metrics include accuracy, AUC, PR-AUC, precision, recall, and F1-score. Among the six models evaluated, Logistic Regression has the highest scores across all metrics, achieving an accuracy score of 0.9995, an AUC score of 1.0000, a PR-AUC score of 1.0000, a precision score of 0.9990, a recall score of 1.0000, and an F1-score of 0.9995. This indicates that Logistic Regression is highly effective in pricing American options within the in-time sample, showing exceptional performance without apparent overfitting or class imbalance issues.

\vspace{0.5cm}

\begin{table}[H]
    \centering
    \begin{tabular}{|c|c|c|c|c|c|c|}
        \hline
        \textbf{Model} & \textbf{Accuracy} & \textbf{AUC} & \textbf{PR-AUC} & \textbf{Precision} & \textbf{Recall} & \textbf{F1-Score} \\
        \hline
        KNN & 0.5101 & 0.6677 & 0.6588 & 0.5090 & 1 & 6746 \\
        \hline
        Decision Tree & 0.5599 & 0.6041 &0. 5813 & 5370 & 0.9663 & 0.6904 \\
        \hline
        XGBoost & 0.5078 & 0.8416 & 0.8425 & 0.5078 & 1 & 0.6736 \\
        \hline
        LightGBM & 0.5078 & 0.864 & 0.5078 & 0.487 & 1 & 0.673 \\
        \hline
        Logistic Regression & 0.9995 & 1 & 1 & 0.999 & 1 & 0.9995 \\
        \hline
        Random Forest & 0.5111 &0.6332 &  0.6174 & 0.5095 & 1 & 0.6750 \\
        \hline
    \end{tabular} 
      \caption{Model evaluation indicators for the in-time sample}
      \label{t11}
\end{table}

LightGBM follows Logistic Regression to have better performance with an AUC score of 0.8640. However, its accuracy score of 0.5078 and PR-AUC score of 0.5078, along with a precision score of 0.4870 and a perfect recall score of 1.0000, indicate a trade-off between precision and recall. This suggests that while LightGBM is good at identifying positive cases, it may produce more false positives compared to Logistic Regression. XGBoost provides robust performance with an AUC score of 0.8416 and a PR-AUC score of 0.8425. It has an accuracy score of 0.5078 and a precision score of 0.5078, coupled with a perfect recall score of 1.0000. The lower precision score compared to its recall indicates a higher likelihood of false positives, but its overall high AUC and PR-AUC scores reflect strong model reliability. Decision Tree offers a balanced performance with an accuracy score of 0.5599, an AUC score of 0.6041, and a PR-AUC score of 0.5813. It has a precision score of 0.5370 and a recall score of 0.9663, leading to an F1-score of 0.6904. This suggests that Decision Tree captures most positive cases effectively, although it has lower discriminative power compared to LightGBM and XGBoost. Random Forest shows moderate performance with an accuracy score of 0.5111, an AUC score of 0.6332, and a PR-AUC score of 0.6174. Its precision score of 0.5095 and recall score of 1.0000 lead to an F1-score of 0.6750. Like the Decision Tree, Random Forest provides balanced but lower overall performance compared to the boosting models. KNN also provides moderate performance with an accuracy score of 0.5101, an AUC score of 0.6677, and a PR-AUC score of 0.6588. It has a precision score of 0.5090 and a recall score of 1.0000, resulting in an F1-score of 0.6746. KNN's performance is similar to Random Forest, with high recall and moderate precision, making it suitable for simpler use cases. Overall, while Logistic Regression is the top performer across all metrics, other models like LightGBM and XGBoost also demonstrate effectiveness, particularly in terms of AUC and PR-AUC, highlighting their robustness in pricing American options.

%      \begin{figure}[b]{0.5\textwidth}
%          \centering
% \includegraphics[width=1\textwidth]{AIMS_Rwanda-Research-Project_Template2022-23/rr.png}
%     \caption{}
%     \label{roc}
%      \end{figure}
%  %\hfill
%      \begin{figure}[b]{0.5\textwidth}
%          \centering
%          \includegraphics[width=1\textwidth]{AIMS_Rwanda-Research-Project_Template2022-23/pr.png}
%     %\caption{}
%     \label{prr}
%      \end{figure}
%      \caption{(a) ROC-AUC curve for the models. (b)Precision-recall curve for the models.}
%      \label{evaluationmodels}

\subsection{ROC-AUC and Precision-Recall Curve Analysis}

\begin{figure}[H]
    \centering
    \begin{minipage}{0.49\textwidth}
        \centering
        \includegraphics[width=\linewidth]{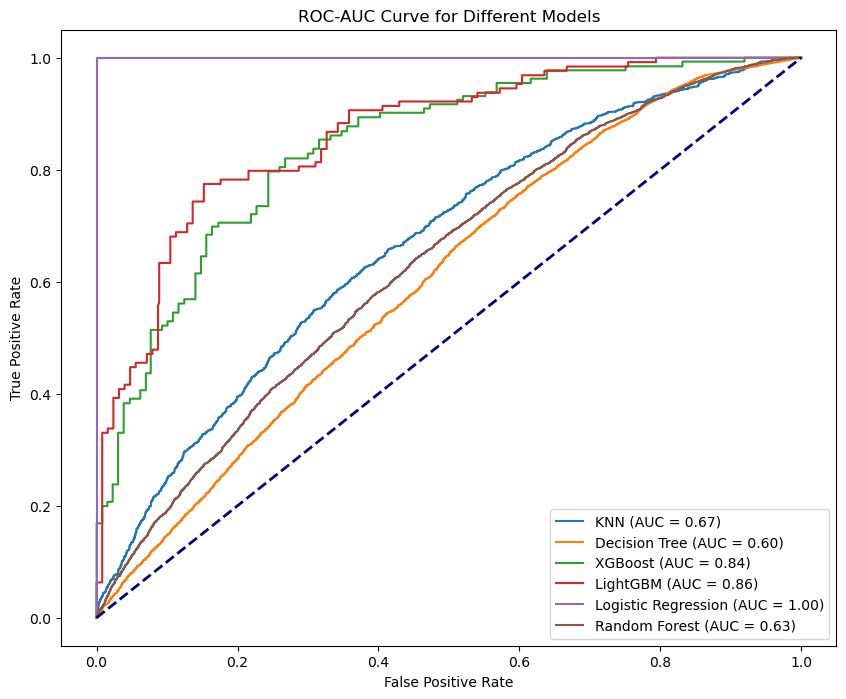}
        \caption{ROC-AUC curve for the models}
        \label{roc}
    \end{minipage}
    \hfill
    \begin{minipage}{0.49\textwidth}
        \centering
        \includegraphics[width=\linewidth]{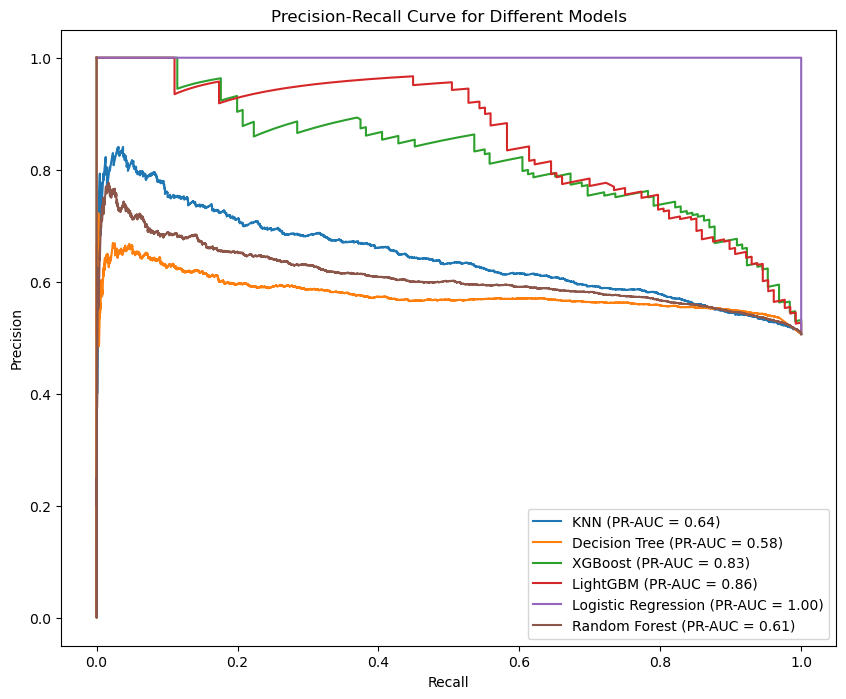}
        \caption{Precision-recall curve for the models}
        \label{pr}
    \end{minipage}
\end{figure}

 Figure \ref{roc} shows the ROC-AUC curve which is depicting the ability of each model to distinguish between positive and negative samples. From the curve, it is evident that:

\begin{itemize}
    \item Logistic Regression has the highest AUC score of 1.00, indicating perfect discriminatory ability.
    \item LightGBM follows with an AUC score of 0.86.
    \item XGBoost has an AUC score of 0.84.
    \item KNN and Random Forest have moderate AUC scores of 0.67 and 0.63, respectively.
    \item The Decision Tree model has the lowest AUC score of 0.60.
\end{itemize}

An AUC score above 0.8 generally indicates a good model, and hence, Logistic Regression, LightGBM, and XGBoost are considered better performers in distinguishing between classes.

% \begin{figure}[H]
%     \centering
%     \includegraphics[width=0.6\textwidth]{pr.png}
%     \caption{Precision-recall curve for the models}
%     \label{pr}
% \end{figure}

Figure \ref{pr} shows the Precision-Recall (PR) curve  
 which provides insights into the trade-off between Precision and Recall for each model. The PR-AUC scores are as follows:
\begin{itemize}
    \item Logistic Regression again leads with a perfect PR-AUC score of 1.00.
    \item LightGBM has a PR-AUC score of 0.86.
    \item XGBoost follows closely with a PR-AUC score of 0.83.
    \item KNN, Random Forest, and Decision Tree have lower PR-AUC scores of 0.64, 0.61, and 0.58, respectively.
\end{itemize}
The higher the PR-AUC score, the better the model is at balancing precision and recall, particularly useful for imbalanced datasets.

\subsection{Confusion Matrix Analysis}

The confusion matrices in Figure \ref{figrrrr} provide a detailed breakdown of True Positives (TP), True Negatives (TN), False Positives (FP), and False Negatives (FN) for each model. From these matrices:
\begin{itemize}
    \item Logistic Regression achieves nearly perfect classification with minimal misclassifications.
    \item LightGBM and XGBoost show a good balance with high TP and TN counts.
    \item Decision Tree and Random Forest exhibit more misclassifications compared to the top performers.
    \item KNN shows the highest number of misclassifications, indicating it may not be as effective for this task.
\end{itemize}
\vspace{0.5cm}

\begin{figure}[H]
    \begin{minipage}[b]{0.4\textwidth}
        \centering
        \includegraphics[width=0.8\linewidth]{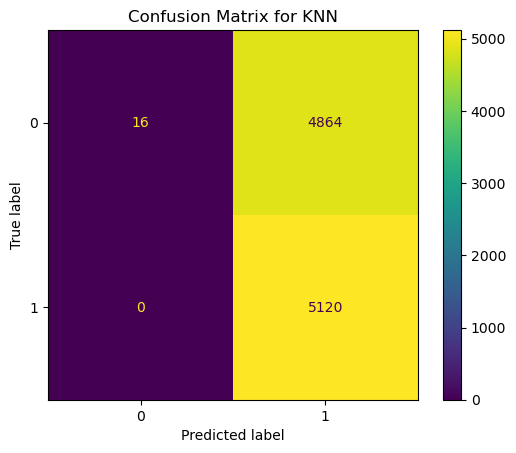}
        %\caption{Confusion Matrix For KNN Model}
        \label{figyyy}
    \end{minipage}
    \hfill
    \begin{minipage}[b]{0.4\textwidth}
        \centering
         \includegraphics[width=0.8\linewidth]{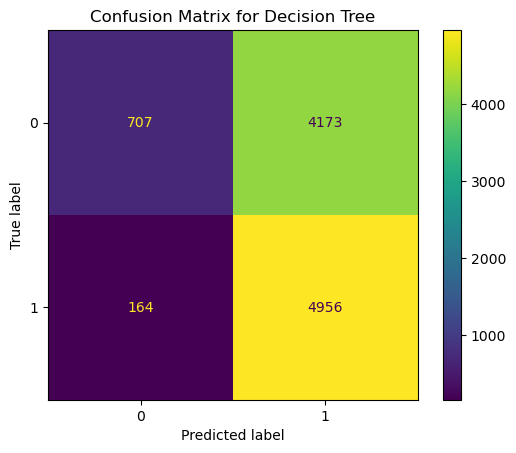}
         %\caption{ROC For KNN Model.}
       \label{figuuyy}
    \end{minipage}
    %\caption{Impact for different values of \(\alpha\) and \(\omega_1\)  on susceptible and TB-infected population.}
    \label{figrrrr}
\end{figure}

\begin{figure}[H]
    \begin{minipage}[b]{0.4\textwidth}
        \centering
       \includegraphics[width=0.8\linewidth]{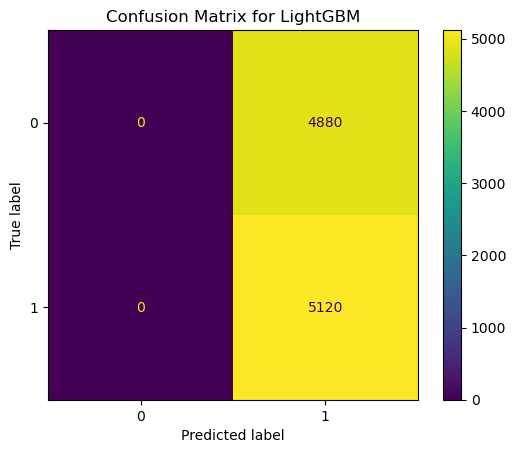}
        %\caption{Confusion Matrix forDecision Tree}
        \label{figyyy}
    \end{minipage}
    \hfill
    \begin{minipage}[b]{0.4\textwidth}
        \centering
        \includegraphics[width=0.8\linewidth]{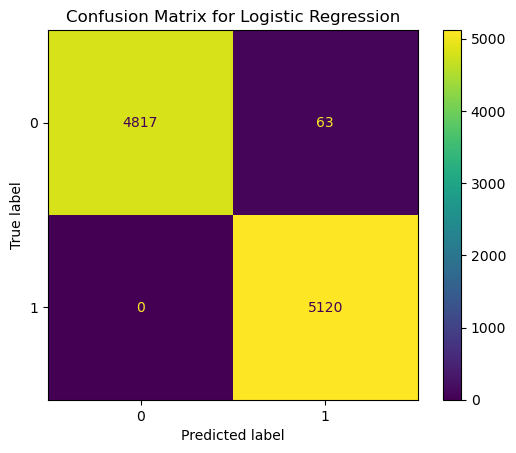}
         %\caption{ROC for Decision Tree.}
       \label{figuuyy}
    \end{minipage}
    %\caption{Impact for different values of \(\alpha\) and \(\omega_1\)  on susceptible and TB-infected population.}
    \label{figrrrr}
\end{figure}

\begin{figure}[H]
    \begin{minipage}[b]{0.4\textwidth}
        \centering
       \includegraphics[width=0.8\linewidth]{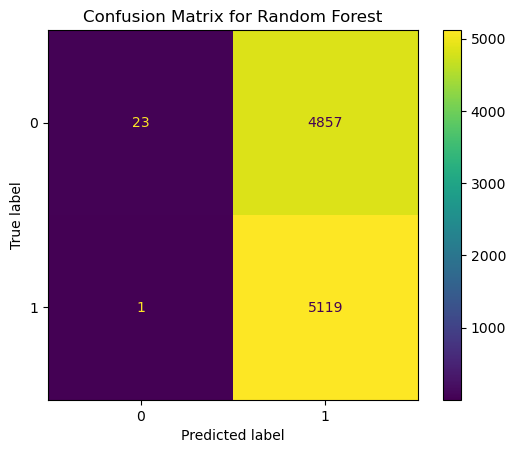}
        %\caption{Confusion Matrix forDecision Tree}
        \label{figyyy}
    \end{minipage}
    \hfill
    \begin{minipage}[b]{0.4\textwidth}
        \centering
        \includegraphics[width=0.8\linewidth]{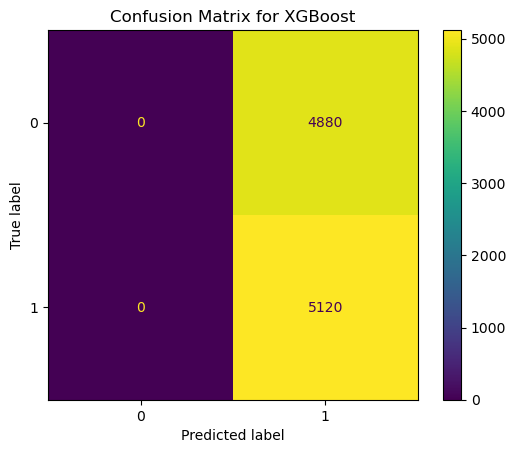}
         %\caption{ROC for Decision Tree.}
       \label{figuuyy}
    \end{minipage}
    \caption{ Confusion matrices of KNN, XGBoost, LightGBM, Logistic
Regression Models,  and Decision Tree.}
    \label{figrrrr}
\end{figure}

Overall, Logistic Regression emerges as the top-performing model across most metrics, making it highly suitable for pricing American options. LightGBM and XGBoost also demonstrate strong performance, while KNN and Decision Tree are less effective.

\subsection{Result of Recurrent Neural Network (RNN) on Dataset}
%\subsection{Data Preparation}
The dataset containing various features related to financial options was loaded and split into features (\textit{X}) and the target variable (\textit{y}), representing bid prices. A train-validation split of 80-20 was employed to partition the dataset into training and validation sets.
\subsection{Model Architecture}

Using TensorFlow's Keras API, LSTM and GRU models were constructed with four hidden fully connected layers, each comprising 200 neurons, followed by an output layer with a single neuron for bid price prediction. Rectified Linear Unit (ReLU) activation functions were applied in all layers. The mean squared error (MSE) loss function was chosen to quantify prediction errors. Both models were optimized using the Adam optimizer with a learning rate of 0.001.
\vspace{0.5cm}

%\section{Results}

% %\subsection{Hyperparameter Comparison (\ref{t1})}
% \begin{table}[H]
%     \centering
%     \caption{Hyperparameters for LSTM and GRU models}
%     \begin{tabular}{|c|c|c|}
%     \toprule
%     Hyperparameter & LSTM                    & GRU                     \\ \hline
%     Activation     & ReLU                    & ReLU                    \\  \hline
%     Loss Function  & MSE                     & MSE                     \\  \hline
%     Neurons        & [200, 200, 200, 200, 1] & [200, 200, 200, 200, 1] \\  \hline
%     Learning Rate  & 0.001                   & 0.001                   \\  \hline
%     Optimizer      & Adam       &Adam \\ 
%     \bottomrule
%     \label{t1}
%     \end{tabular}
% \end{table}
\begin{table}[H]
    \centering
    \label{t1}
    \begin{tabular}{|c|c|c|}
    \hline
    Hyperparameter & LSTM                    & GRU                     \\ \hline
    Activation     & ReLU                    & ReLU                    \\ \hline
    Loss Function  & MSE                     & MSE                     \\ \hline
    Neurons        & [200, 200, 200, 200, 1] & [200, 200, 200, 200, 1] \\ \hline
    Learning Rate  & 0.001                   & 0.001                   \\ \hline
    Optimizer      & Adam                    & Adam                    \\ 
    \hline 
    \end{tabular}
    \caption{Hyperparameters for LSTM and GRU models}
    \label{trb}
\end{table}

The models underwent training for 200 epochs with a batch size of 64. During training, validation set performance was monitored to prevent over fitting. Evaluation of the models was conducted using the mean squared error (MSE) metric on the validation set to assess predictive accuracy which are summarised in Table \ref{trb}.

% \subsection{Pricing Performance Comparison}
% To evaluate the pricing performance, we compared the deep learning models (LSTM and GRU) with Least Squares Monte Carlo (LSM) method for pricing American options. Performance assessment was based on pricing errors for American options using metrics such as MSE, RMSE, and MAE. Our results were summarized and compared with those obtained from the LSM method to ascertain the effectiveness of deep learning models in option pricing.

% \subsection{Visualization}
% Visualization of the training and validation loss curves for both LSTM and GRU models was performed to observe their learning behavior and identify potential overfitting or underfitting issues. 
% Through our experimentation, we aimed to validate the effectiveness of LSTM and GRU models in predicting bid prices for financial options. By comparing their performance with the LSM method for pricing American options, we sought to understand the strengths and limitations of deep learning approaches in option pricing. Our results, coupled with comprehensive analysis, provide valuable insights into the application of deep learning in financial forecasting tasks.

\subsection{Training and Validation Loss Analysis}

\begin{figure}[H]
    \centering
    \begin{minipage}{0.49\textwidth}
        \centering
        \includegraphics[width=\linewidth]{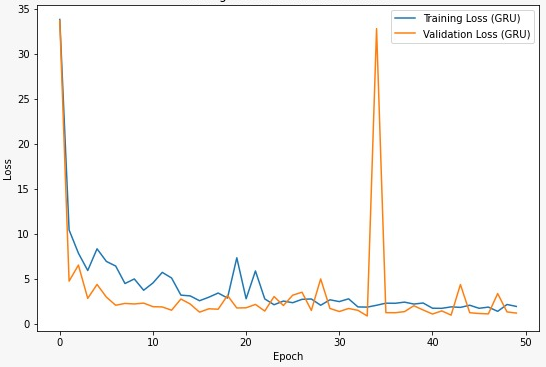}
        \caption{Train and test loss of LSTM model}
        \label{eva}
    \end{minipage}
    \hfill
    \begin{minipage}{0.49\textwidth}
        \centering
        \includegraphics[width=\linewidth]{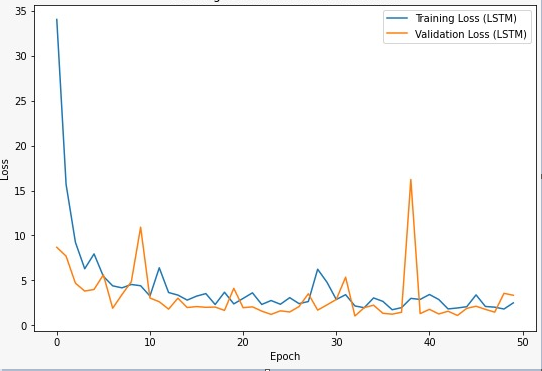}
        \caption{Train and test loss of LSTM model}
        \label{evaa}
    \end{minipage}
\end{figure}

From the training and validation loss Figure \ref{eva} and \ref{evaa}, it is evident that both the GRU and LSTM models demonstrate significant reductions in loss over the training epochs. The GRU model shows more stability with fewer spikes in the validation loss compared to the LSTM model, which has a noticeable spike around the 35th epoch.

 The GRU model's validation loss fluctuates less and maintains a relatively lower loss towards the end of the training, suggesting better generalization and stability. On the other hand, the LSTM model, while initially reducing loss effectively, exhibits more variability and occasional higher spikes, indicating potential overfitting or sensitivity to certain epochs.

% \begin{figure}[H]
%      \centering
%      \begin{subfigure}[b]{0.46\textwidth}
%          \centering
% \includegraphics[width=0.8\textwidth]{Ca11.PNG}
%     \caption{}
%     \label{fig:gru_loss}
%      \end{subfigure}
%  \hfill
%      \begin{subfigure}[b]{0.46\textwidth}
%          \centering
%           \includegraphics[width=0.8\textwidth]{Caaa1.PNG}
%     \caption{}
%     \label{fig:lstm_loss}
%     \label{pr}
%      \end{subfigure}
%      \caption{ GRU model train and test loss (a) and LSTM model train and test loss (b).}
%      \label{eva}
%      \end{figure}

% \subsection{Pricing Performance Comparison}
% To evaluate the pricing performance, we compared the deep learning models (LSTM and GRU) with Least Squares Monte Carlo (LSM) method for pricing American options. Performance assessment was based on pricing errors for American options using metrics such as MSE, RMSE, and MAE. Our results were summarized and compared with those obtained from the LSM method to ascertain the effectiveness of deep learning models in option pricing.

\subsection{Error Metrics Comparison}
Looking at the error metrics provided in Table \ref{epo}, the GRU model outperforms the LSTM model across all evaluated metrics. The GRU model achieves a lower Mean Absolute Error (MAE) of 0.49075 compared to 0.5919 for the LSTM model.

\vspace{0.5cm}

\begin{table}[H]
    \centering
    \begin{tabular}{|l|l|l|l|l|l|l|l|}
    \hline
    Options Type & Model & Train/Test (\%) & Epochs & Time & MAE & MSE & RMSE \\ \hline
    
    Call & LSTM & 80/20 & 200 & 46s  7ms/step & 0.5919 & 1.7017 & 1.3045\\ \hline
    Call & GRU & 80/20 & 200 & 36s  6ms/step & 0.49075 &  0.84277 & 0.9180 \\ \hline 
    \end{tabular} 
    \caption{Deep learning error metrics }
    \label{epo}
\end{table}

 Similarly, the Mean Squared Error (MSE) and Root Mean Squared Error (RMSE) are significantly lower for the GRU model, with values of 0.84277 and 0.9180 respectively, as opposed to the LSTM model's 1.7017 and 1.3045. This indicates that the GRU model not only fits the training data better but also predicts more accurately on the test data, making it a more reliable choice for this specific application. Additionally, the training time per epoch is shorter for the GRU model (6ms/step) compared to the LSTM model (7ms/step), indicating a more efficient training process. Therefore, the GRU model demonstrates superior performance in terms of both stability and predictive accuracy, making it a preferable choice over the LSTM model.

\section{Conclusions}

The study demonstrates that machine learning algorithms integrated with Monte Carlo simulations can effectively price American options, offering significant improvements over traditional methods. Through extensive experimentation, we found that models such as neural networks and other machine learning techniques provide more accurate pricing, especially in complex market conditions where traditional models struggle. Moreover, the study demonstrates the effectiveness of LSTM and GRU models in predicting bid prices for financial options, with the GRU model exhibiting superior performance. These results pave the way for further exploration and optimization of deep learning methods in financial forecasting, suggesting that such models can potentially outperform traditional approaches like LSM in specific applications. The comprehensive analysis and comparison provide valuable insights into the strengths and limitations of deep learning models, advocating for their integration into financial forecasting and option pricing frameworks. The findings suggest several future directions for research. One promising area is the integration of more sophisticated machine learning models, such as deep reinforcement learning and advanced neural network architectures, to further enhance pricing accuracy. Additionally, the development of hybrid models that combine the strengths of traditional financial theories and machine learning could provide even more robust solutions. Future work should also focus on improving model interpretability and addressing data scarcity issues by leveraging techniques such as transfer learning and data augmentation. Finally, real-world testing and validation of these models in live trading environments will be crucial for assessing their practical applicability and robustness. Again future research could focus on optimizing hyperparameters, exploring other deep learning architectures like Transformers, and incorporating additional features such as macroeconomic indicators. Data augmentation and synthetic data generation techniques can enhance model training, while real-time prediction systems would validate practical performance. By advancing the intersection of machine learning and financial modeling, this research opens up new possibilities for more accurate and efficient option pricing, ultimately contributing to better risk management and decision-making in financial markets.

		\bibliography{references}
		
	\end{document}